\documentclass[10pt,twocolumn,letterpaper]{article}

\usepackage{cvpr}              

\usepackage{graphicx}
\usepackage{amsmath}
\usepackage{amssymb}
\usepackage{booktabs}
\usepackage{enumitem}
\usepackage[linesnumbered,ruled,vlined]{algorithm2e}
\usepackage{tabularray} 
\usepackage{hhline} 
\usepackage{arydshln}

\usepackage[pagebackref,breaklinks,colorlinks]{hyperref}

\usepackage[capitalize]{cleveref}
\crefname{section}{Sec.}{Secs.}
\Crefname{section}{Section}{Sections}
\Crefname{table}{Table}{Tables}
\crefname{table}{Tab.}{Tabs.}

\begin{document}

\title{FODVid: Flow-guided Object Discovery in Videos}
\author{Silky Singh~~~~~~~Shripad Deshmukh~~~~~~~Mausoom Sarkar~~~~~~~Rishabh Jain\\
~~~~~~Mayur Hemani~~~~~~~Balaji Krishnamurthy\\
Media and Data Science Research, Adobe\\
{\tt\small \{silsingh,~shdeshmu,~msarkar,~rishabhj,~mayur,~kbalaji\}@adobe.com}
}
\maketitle


\begin{abstract}
    Segmentation of objects in a video is challenging due to the nuances such as motion blurring, parallax, occlusions, changes in illumination, etc. Instead of addressing these nuances separately, we focus on building a generalizable solution that avoids overfitting to the individual intricacies. Such a solution would also help us save enormous resources involved in human annotation of video corpora. To solve Video Object Segmentation (VOS) in an unsupervised setting, we propose a new pipeline (\textbf{FODVid}) based on the idea of guiding segmentation outputs using flow-guided graph-cut and temporal consistency. Basically, we design a segmentation model incorporating intra-frame appearance and flow similarities, and inter-frame temporal continuation of the objects under consideration. We perform an extensive experimental analysis of our straightforward methodology on the standard DAVIS16 video benchmark. Though simple, our approach produces results comparable (within a range of $\sim 2$ mIoU) to the existing top approaches in unsupervised VOS. The simplicity and effectiveness of our technique opens up new avenues for research in the video domain.
\end{abstract}

\vspace{-0.5cm}

\section{Introduction}
\label{sec:introduction}

\begin{figure*}[h]
\begin{center}
\includegraphics[width=0.8\linewidth]{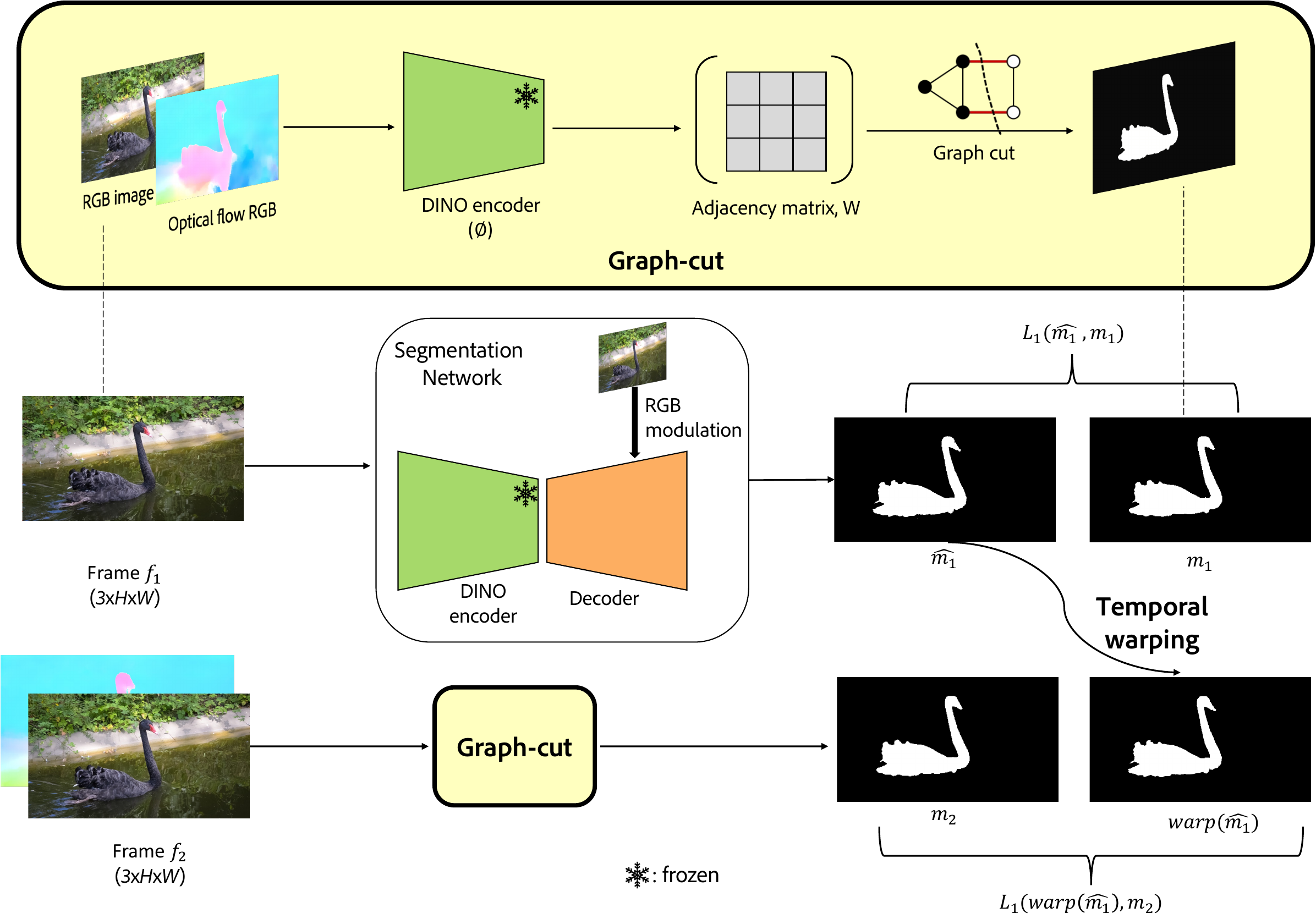}
\end{center}
\caption{\textbf{FODVid: Our proposed pipeline for unsupervised video object segmentation.} An image and its corresponding optical flow RGB are featurized using DINO~\cite{caron2021emerging} and graph-cut is performed to produce a set of preliminary object segmentation masks. These masks are then used as pseudo ground-truths to train a segmentation network by enforcing temporal consistency between nearby frames.}
\label{fig:model_architecture}
\end{figure*}

Object segmentation, in its various forms, is a widely studied problem in computer vision~\cite{long2015fully}. The classic task finds critical applications across multiple domains, such as autonomous driving, augmented reality, human-computer interaction, video summarization etc. The deep neural networks employed for this purpose are typically trained on large annotated datasets created through enormous human efforts that take several months of focused work. Moreover, training a segmentation model on one such dataset does not guarantee transfer-ability to real-world data since dataset-specific considerations in model design overfit the model to a particular use case. These issues call attention to developing generalizable solutions that can work with minimal human supervision.

To alleviate some of the problems with supervised segmentation, methods that work on weaker forms of human supervision were proposed. These methods function with weak supervision provided through scribbles~\cite{lin2016scribblesup, tang2018normalized, tang2018regularized, vernaza2017learning, xu2015learning} or clicks~\cite{bearman2016s} or image-level tags~\cite{papandreou2015weakly, tang2018regularized, xu2015learning} or even bounds~\cite{dai2015boxsup}. Further, semi-supervised segmentation techniques~\cite{dai2015boxsup, hong2015decoupled, hung2018adversarial, papandreou2015weakly, pathak2015constrained} were proposed that attempt segmentation in a setting where only a fraction of the image datasets are human-labelled. Nonetheless, both weak-supervised and semi-supervised techniques still rely on bulky human supervision directly or indirectly. Therefore, in the present work, we focus on segmenting objects in a video without relying on any form of external supervision.


The aim of Video Object Segmentation (VOS) is to localize the most salient object(s) in a given video frame~\cite{perazzi2016benchmark}. In the literature~\cite{li2013video, ochs2013segmentation, perazzi2016benchmark, xie2022segmenting, ye2022deformable}, VOS is generally re-framed as a foreground-background separation problem. We propose an end-to-end pipeline as follows -- for a given video frame, we encode its frames and their optical flows in RGB format using a self-supervised ViT like DINO~\cite{caron2021emerging}. The image and flow features are then used in a linear combination to form a similarity based adjacency matrix between the frame patches. By performing graph-cut on this adjacency matrix, we obtain a preliminary set of segmentation masks for all the frames in a video. These masks are then used as pseudo-ground truths to train a segmentation network. We devise a loss schedule that alternates between graph-cut mask of current frame and a nearby frame to enforce temporal consistency while training the segmentation network(refer \ref{sec:temporal_warping}).

We summarize the contributions of this work below:


\begin{enumerate}[nosep]

    \item We present a new pipeline (FODVid) for unsupervised Video Object Segmentation (VOS), utilising both appearance and motion information contained in videos.
    
    
    \item We demonstrate the importance of perceptual motion cues for object discovery. In particular, we employ the Gestalt principle, ``things that move together, belong together" to enrich frame patch similarity.
    
    \item Our methodology is simple to implement and produces results comparable to existing top unsupervised VOS approaches. We achieve an mIoU score of $\textbf{78.71}$ on the standard DAVIS16 benchmark. Additionally, the proposed temporal refinement provides an improvement of as much as $\textbf{+9.88 mIoU}$ on certain video sequences.
    
    


    
\end{enumerate}

\section{Methodology}
\label{sec:methodology}

\begin{figure*}[h]
\begin{center}
\includegraphics[width=0.8\linewidth]{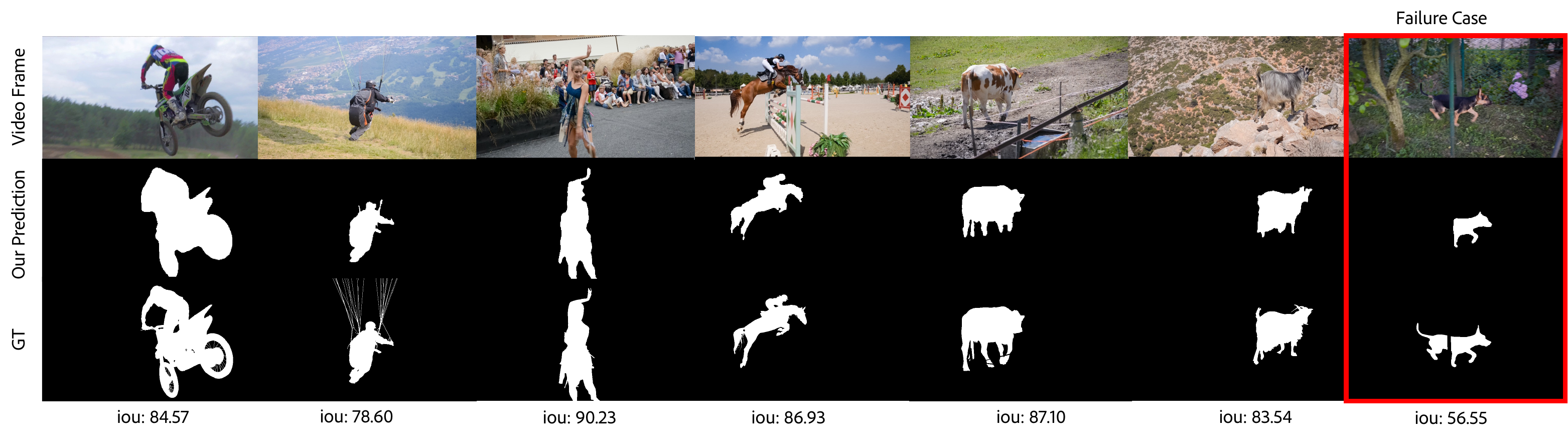}
\end{center}
   \caption{\textbf{Qualitative results of our method on DAVIS16 dataset}. Top row: RGB video frame, middle row: our prediction, bottom row: ground truth. The last column shows a failure case where only a part of the occluded object is identified.}
\label{fig:qualitative_results}
\end{figure*}

Our approach involves guiding segmentation through -- 1) graph-cut and 2)  temporal warping via optical flow (Refer Fig.~\ref{fig:model_architecture} for our proposed FODVid pipeline). 

\vspace{-0.1cm}
\subsection{Graph-cut}
\label{sec:graph_cut}


Consider video frame $f$, objects of which we wish to segment out. We start by creating a fully-connected graph $G= (V, E)$, where $V$ denotes the set of vertices obtained by dividing $f$ into square patches of size $p_s\times p_s$, and $E$ denotes the set of edges such that each edge weight quantifies the similarity between connecting vertices (in our case, image patches). Formally, the adjacency matrix $W$ underlying $G$ is made of $w_{ij} = S(v_i, v_j)$, where $S(\cdot)$ denotes the similarity measure between two given vertices (patches).




When compared to standalone images, video frames are special as they track information about a set of objects temporally, in continuation. Since the main aim of VOS is to segment out such objects, we wish to incorporate perceptual nuances in the similarity measure $S$. Based on the Gestalt principle of common fate, image patches having similar flow direction and magnitude most likely belong to the same object. We utilise this complementary motion information while defining $W$.

 Formally, the overall similarity score between two patches from the same frame is defined as a linear combination of similarity between standard patch embeddings (obtained using DINO encoder~\cite{caron2021emerging}) and that between DINO embeddings of the RGB-optical flow at the respective patches.  In mathematical notation,
\[
S(v_i, v_j) = \alpha \cdot S'(\phi(v_i), \phi(v_j)) + (1 - \alpha) \cdot S'(\phi(\psi(v_i)), \phi(\psi(v_j)))
\]

where $\alpha \in [0,1]$, $\phi(\cdot)$ denotes the DINO encoder, $\psi(\cdot)$ denotes the RGB-optical flow estimator, i.e., model computing optical flow in 3-channel RGB image format, and $S'(\cdot)$ is the cosine similarity function, given by $S'(\Vec{x}, \Vec{y}) = \frac{\Vec{x}\cdot \Vec{y}}{||\Vec{x}||_2 ||\Vec{y}||_2}$. We obtain $W$ using the $S$ defined above.




Further, in order to minimize the variance, we normalize $w_{ij}$'s by thresholding them.

$$
w_{ij} \xleftarrow{} \begin{cases}
			1, & \text{if $w_{ij}$ $\geq \tau$} \\
            \epsilon, & \text{otherwise}
		 \end{cases}
$$
where $\tau$ denotes the weight threshold hyper-parameter, and value of $\epsilon$ is set to $10^{-5} (\neq 0)$ to ensure the fully connectedness of $G$. 

The solution to graph-cut is well studied in the literature~\cite{wang2022self, wang2023cut, shi2000normalized}. We compute the second-smallest eigenvector of the matrix $W$, and threshold on the average value to create a bi-partition of $G$; the partitions representing foreground and background. We use these binary masks as pseudo-ground truths for training an encoder-decoder style segmentation network.


\subsection{Temporal warping via optical flow}
\label{sec:temporal_warping}

The graph-cut approach described in the previous section relies only on information from the single frame under consideration. As a side note, we did not observe any gains in performance of the segmentation network by direct distillation using graph-cut masks- motivating us to devise this video-level temporal refinement scheme. Moreover, the graph-cut masks produced in Sec.~\ref{sec:graph_cut} are noisy -- part of background object is identified as foreground or vice-versa. As such, we intuit that the extra information needed to discern foreground from background can come from nearby frames in the temporal neighborhood of the current frame.


Let the frame under consideration be $f_1$ and the graph-cut mask obtained per the procedure described in Sec.~\ref{sec:graph_cut} be $m_1$. We sample frame $f_2$ in the \{-2, -1, +1, +2\} temporal neighbourhood of $f_1$. Let $m_2$ denote the graph-cut mask for $f_2$. Let the prediction of the segmentation network for $f_1$ be $\hat{m_1}$.

We design a loss schedule for training the segmentation network such that, for 50\% of the time, we use segmentation-loss between $m_1$ and $\hat{m_1}$. For the remaining 50\% of the time, we temporally warp $\hat{m_1}$ using optical flow estimated between $f_1$ and $f_2$ to obtain segmentation mask prediction for $f_2$ and take the segmentation loss between this mask and $m_2$. Under the assumption that all pseudo-ground truths are equally noisy, we give equal consideration to both the branches~(note the threshold of 0.5 for $p$ in Alg.~\ref{alg:loss_schedule}). The algorithmic description is provided in Alg.~\ref{alg:loss_schedule}.



\begin{algorithm}
\footnotesize
\caption{Loss Schedule}
\label{alg:loss_schedule}
\SetKwInOut{Input}{Input}
\SetKwInOut{Initialize}{Initialize}
\SetKwInOut{Output}{Output}
\vspace{1mm}

\For{epoch in $\{1, 2, ..., N\}$}{
    $p \sim \mathcal{U}(0, 1)$ \tcp{sample from uniform(0,1)}
    \uIf{$p < 0.5$}{
        $L = || \hat{m_1} - m_1 ||_1$ \tcp{graph-cut guidance}
    }
    \Else{
        $L = || \text{warp}(\hat{m_1}) - m_2 ||_1$  \tcp{enforcing temporal consistency}
    }
    Update weights based on the computed $L$ 
}
\end{algorithm}

\section{Experiments and Results}
\label{sec:experiments}

\vspace{-0.1cm}
\subsection{Experimental Setup}
We train the segmentation network using 4 NVIDIA A100 80GB GPUs for 200 epochs with a batch size of 16. We employ Adam optimizer~\cite{kingma2014adam} with a learning rate of $10^{-4}$, momentum terms set to $\beta_1 = 0.9$ and $\beta_2 = 0.999$. The training is performed at an image resolution of $256\times512$. We borrow the decoder architecture from SPADE~\cite{park2019semantic} and keep DINO~\cite{caron2021emerging} as the encoder. We utilise optical flow obtained using pre-trained ARFlow~\cite{liu2020learning} in the graph-cut stage~(Sec.~\ref{sec:graph_cut}), and pre-trained GMFlow~\cite{xu2022gmflow} for warping the segmentation masks in the second step~(Sec.~\ref{sec:temporal_warping}). 





\vspace{-0.1cm}
\subsection{Inference Setup}
One of the key advantages of our method is that the resultant segmentation network does not require optical flow as input unlike many VOS approaches. The segmentation network, while trained on videos, can work solely on images. As a result, we can infer the object segmentation for all the frames in a video independently.


\vspace{-0.1cm}
\subsection{Results}

Table~\ref{tab:quantitative_comparisons} depicts the comparison between our method and the existing unsupervised VOS approaches. Further, in tables~\ref{tab:dino_ablations} and~\ref{tab:hyperparam_ablations}, we depict the ablations on DINO architecture and the important hyper-parameters $\tau$ (used for edge thresholding) and $\alpha$ (used for similarity combination). Moreover, we perform an ablation on the flow estimator used in graph-cut experiments and present its results in table~\ref{tab:flow_model_ablation}.

\begin{figure}[htp]
    \centering
    \includegraphics[width=1.0\linewidth]{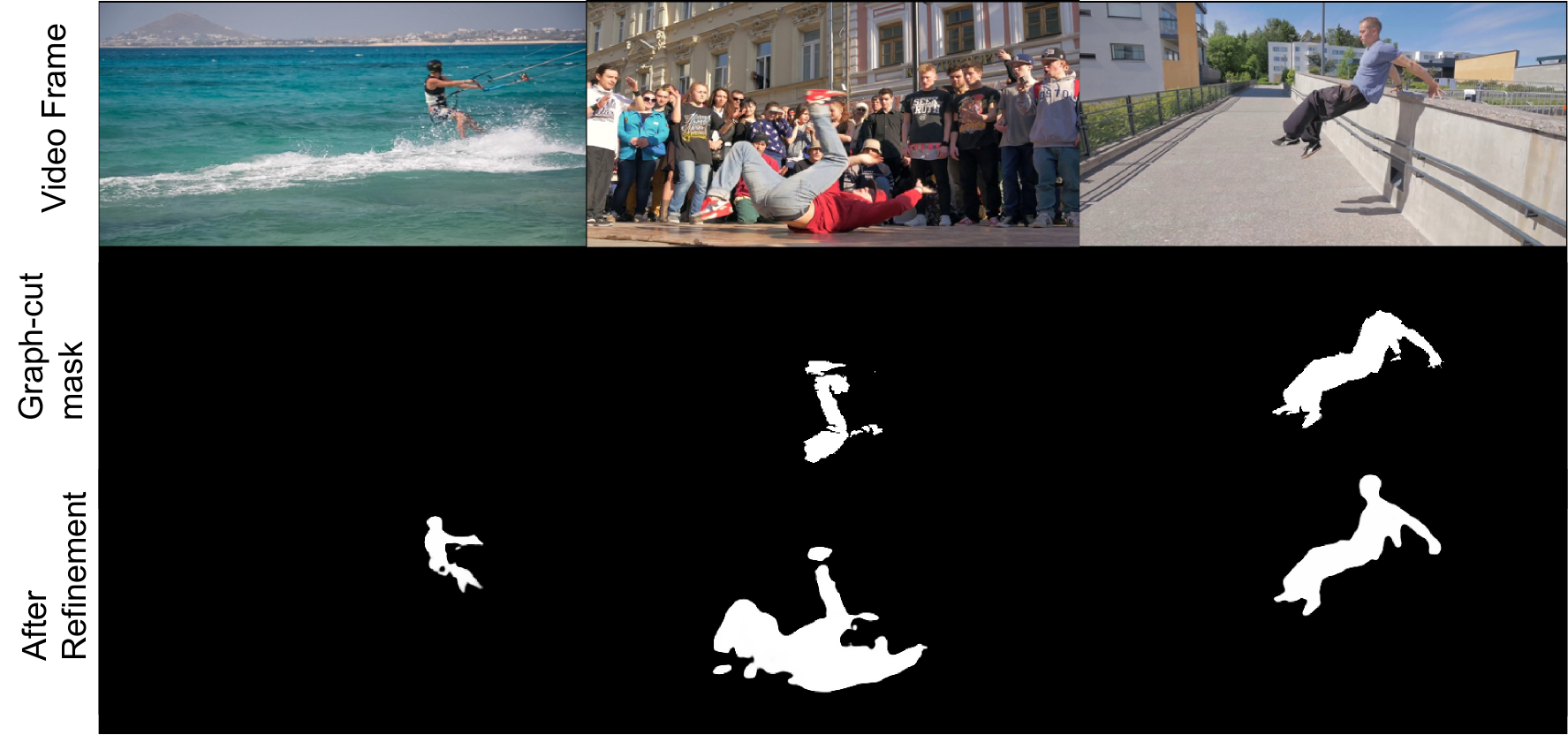}
    \caption{\textbf{Improvements from guidance through temporal warping} (Refer Sec.~\ref{sec:temporal_warping}). Top row: image, middle row: graph-cut masks, bottom row: after temporal refinement. We observe that temporal warping enables the model to predict an object when it is absent in the graph-cut mask. As shown, it generates a person's whole body while only parts of it are captured through graph-cut.}
    \label{fig:temporal_warping_improvements}
\end{figure}

We also show the qualitative results of our method on examples from the DAVIS16 dataset in figure~\ref{fig:qualitative_results}. Additionally, we demonstrate the qualitative improvements observed because of the temporal refinement in figure~\ref{fig:temporal_warping_improvements}.

\begin{table}[]
\begin{center}
\begin{tabular}{|l|l|c|}
\hline
\textbf{Method} & \textbf{Flow Method} & \textbf{mIoU($J \uparrow$)} \\
\hline
SAGE~\cite{wang2017saliency} & LDOF~\cite{brox2010large} & 42.6 \\
NLC~\cite{faktor2014video} & SIFTFlow~\cite{liu2009beyond} & 55.1 \\
CUT~\cite{keuper2015motion} & LDOF~\cite{brox2010large} & 55.2 \\
FTS~\cite{papazoglou2013fast} & LDOF~\cite{brox2010large} &  55.8 \\
\hdashline
AMD~\cite{liu2021emergence} & $\times$ & 57.8 \\
MG~\cite{yang2021self} & RAFT~\cite{teed2020raft} & 68.3 \\
EM~\cite{meunier2022driven} & RAFT~\cite{teed2020raft} &  69.3 \\
CIS $^{*}$~\cite{yang2019unsupervised} & PWCNet~\cite{sun2018pwc} & 71.5 \\
ARP $^{**}$~\cite{koh2017primary} & CPMFlow~\cite{hu2016efficient} & 76.2 \\
OCLR $^{\ddagger}$~\cite{xie2022segmenting} & RAFT~\cite{teed2020raft} & 78.9 \\
DS $^{\dagger}$~\cite{ye2022deformable} & RAFT~\cite{teed2020raft} & 79.1 \\
DyStaB $^{**}$~\cite{yang2021dystab} & RAFT~\cite{teed2020raft} & 80.0 \\
Ponimatkin \etal~\cite{ponimatkin2023simple} & ARFlow~\cite{liu2020learning} & 80.2 \\
Guess What Moves $^{*}$~\cite{choudhury2022guess} & RAFT~\cite{teed2020raft} & \textbf{80.7} \\
\hline
Ours (FODVid) $^{\dagger}$ & ARFlow~\cite{liu2020learning} $\&$ & 78.71 \\
 & GMFlow~\cite{xu2022gmflow} &  \\
\hline
\end{tabular}
\end{center}
\caption{\textbf{Quantitative comparison with unsupervised VOS approaches on the DAVIS16 dataset.} Our segmentation pipeline based on learning-free graph-cut combined with temporal refinement performs favourably to the existing state-of-the-art. The current SOTA mIoU score is shown in bold. (Methods above the dashed line are non deep-learning based approaches.  $^{\dagger}$ denotes optimization per video sequence. $^{**}$ DyStaB utilises supervised pre-training, ARP uses human supervision in the form of saliency maps. ${\ddagger}$ OCLR leverages manual annotations from large-scale Youtube-VOS data to generate synthetic data. $^*$ denotes significant post-processing techniques like multi-step flow, multi-crop ensemble, and temporal smoothing. In contrast, we do not use heavy post-processing techniques or any supervision.)}
\label{tab:quantitative_comparisons}
\end{table}

\begin{table}
\centering
\begin{tabular}{|c|c|c|c|c|}
\hline
\textbf{Architecture}  & \textit{ViT-S/8} & \textit{ViT-S/16}      & \textit{ViT-B/8} & \textit{ViT-B/16}  \\ 
\hhline{|=====|}
\textbf{mIoU($J \uparrow$)} & 73.46            & 74.74       & \textbf{76.76}            & 74.08                    \\
\hline
\end{tabular}
\caption{\textbf{Ablation on the DINO architecture used for Graph-cut} (Refer Sec.~\ref{sec:graph_cut}). We identify ViT-B/8 as the best architecture.}
\label{tab:dino_ablations}
\end{table}

\begin{table}
\centering
\begin{tabular}{|c|c||c|c|} 
\hline
 \textbf{$\tau$} & \textbf{mIoU}  & \textbf{$\alpha$} & \textbf{mIoU}   \\ 
\hline
 0.00                             & 63.05          & 0.2                                & 35.22           \\ 
 0.05                             & 71.32          &  0.3                                & 46.04          \\ 
0.10                             & 74.38          & 0.4                                & 56.63     \\ 
 0.15                             & 75.13          &  0.5                                & 65.20          \\ 
 0.20                             & 75.96          &   0.6                                & 72.55       \\ 
 0.25                             & \textbf{76.76} &   0.7                                & \textbf{75.96}        \\ 
 0.30                             & 76.56          & 0.8                                & 75.31     \\ 
 0.35                             & 75.91          &  0.9                                & 72.19       \\ 
 0.40                             & 75.83          &  1.0 & 62.66         \\
\hline
\end{tabular}
\caption{\textbf{Hyper-parameter ablations.} We find that the similarity edge threshold $\tau = 0.25$ \& the linear combination coefficient $\alpha = 0.7$ gives the best result. Note that $\alpha = 1.0$ corresponds to using only image features for graph-cut. This confirms our hypothesis around combining flow features with raw image features i.e., they improve the quality of segmentation significantly(Refer Sec.~\ref{sec:graph_cut}).}
\label{tab:hyperparam_ablations}
\end{table}



\begin{table}
\centering
\label{tab:ablations_flow_models}
\begin{tblr}{
  cells = {c},
  cell{1}{1} = {r=2}{},
  cell{1}{2} = {c=3}{},
  vlines,
  hline{1,3,5} = {-}{},
  hline{2} = {2-4}{},
}
{\textbf{Flow Model}} & \textbf{mIoU($J \uparrow$) on DAVIS16}           &                          &                   \\
                                & \textit{Graph-cut} & \textit{Post Refinement} & \textit{Post CRF}\\
\textit{GMFlow}~\cite{xu2022gmflow}                 & 76.76        & \textbf{77.94}                    & 76.89             \\
\textit{ARFlow}~\cite{liu2020learning}                 & 78.03       & \textbf{78.71}                    & 77.57                   
\end{tblr}
\caption{\textbf{Ablation on flow model used in Graph-cut.} Interestingly, we find ARFlow, which is trained in a completely unsupervised fashion, to perform better than its supervised alternative. Also, we observe that CRF~\cite{krahenbuhl2011efficient} does not improve the overall quality of obtained segmentation masks.}
\vspace{-0.5cm}
\label{tab:flow_model_ablation}
\end{table}

\vspace{-0.25cm}
\section{Conclusion}
\label{sec:conclusion}

We present a novel solution to unsupervised Video Object Segmentation (VOS). Our straight-forward approach is built on the idea of guiding the segmentation network with pseudo-ground truths from flow-induced graph-cut masks and temporal consistency in videos. We analyse our methodology extensively on the standard DAVIS16 video dataset, where we show the effectiveness of our technique to produce results comparable to existing leading approaches. In the future, we would like to extend this work to other video benchmarks: SegTrackv2~\cite{li2013video} and FBMS59~\cite{ochs2013segmentation}. Finally, through this work, we wish to emphasize the importance of motion cues for object discovery.





{\small
\bibliographystyle{ieee_fullname}
\bibliography{egbib}
}

\end{document}